\documentclass[conference]{IEEEtran}
\IEEEoverridecommandlockouts
\usepackage{amsmath,amsfonts}
\usepackage{algorithmic}
\usepackage{algorithm}
\usepackage{array}
\usepackage{textcomp}
\usepackage{stfloats}
\usepackage{url}
\usepackage{verbatim}
\usepackage{graphicx}
\usepackage[utf8]{inputenc}
\usepackage[T1]{fontenc}
\usepackage{graphicx}
\usepackage{caption}
\usepackage{subcaption}
\usepackage{makecell}

\usepackage{graphics} 
\usepackage{epsfig} 
\usepackage{xcolor}
\usepackage[normalem]{ulem}
\usepackage{booktabs}
\usepackage{multicol}
\usepackage{multirow}
\usepackage{cancel}
\usepackage{cleveref}
\usepackage{comment}
\usepackage{siunitx}

\usepackage[inline]{enumitem}
\usepackage{cite}
\usepackage{amssymb}

\usepackage{acro}
\usepackage{bm}
\usepackage{mathtools}
\usepackage{xcolor}

\DeclareAcronym{ASL}{short = ASL, long = Autonomous Systems Lab}
\DeclareAcronym{OMAV}{short = OMAV, long = Omnidirectional Micro Aerial Vehicle}
\DeclareAcronym{MAV}{short = MAV, long = Micro Aerial Vehicle}
\DeclareAcronym{DoF}{short = DoF, long = degrees of freedom}
\DeclareAcronym{PBC}{short = PBC, long = passivity-based control}
\DeclareAcronym{PH}{short = PH, long = Port-Hamiltonian}
\DeclareAcronym{NDT}{short = NDT, long = non-destructive testing}
\DeclareAcronym{DOE}{short = DOE, long = design of experiments}
\DeclareAcronym{PEMS}{short = PEMS, long = Power and Energy Monitoring System}
\DeclareAcronym{WTC}{short = WTC, long = wrench tracking controller}
\DeclareAcronym{PTC}{short = PTC, long = pose tracking controller}
\DeclareAcronym{MBE}{short = MBE, long = momentum-based wrench estimator}
\DeclareAcronym{ASIC}{short = ASIC, long = Axis-Selective Impedance Control}
\DeclareAcronym{MPC}{short = MPC, long = Model Predictive Control}
\DeclareAcronym{MPPI}{short = MPPI, long = Model Predictive Path Integral}
\DeclareAcronym{APhI}{short = APhI, long = Aerial Physical Interaction}
\DeclareAcronym{LLE}{short = LLE, long = Largest Lyapunov Exponent}
\DeclareAcronym{ICBF}{short = ICBF, long = Integral Control Barrier Function}
\DeclareAcronym{CBF}{short = CBF, long = Control Barrier Function}
\DeclareAcronym{COM}{short = CoM, long = Center of Mass}
\DeclareAcronym{AM}{short = AM, long = Aerial Manipulator}
\DeclareAcronym{MR}{short = MR, long = Mixed Reality}
\DeclareAcronym{AR}{short = AR, long = Augmented Reality}
\DeclareAcronym{VR}{short = VR, long = Virtual Reality}
\DeclareAcronym{HRI}{short = HRI, long = Human-Robot Interaction}
\DeclareAcronym{RL}{short = RL, long = Reinforcement Learning}
\DeclareAcronym{PPO}{short = PPO, long = Proximal Policy Optimization}

\DeclareAcronym{NASA-TLX}{short = NASA-TLX, long = NASA Task Load Index}
\DeclareAcronym{MD}{short = MD, long = mental demand}
\DeclareAcronym{PD}{short = PD, long = physical demand}
\DeclareAcronym{TD}{short = TD, long = temporal demand}
\DeclareAcronym{EF}{short = EF, long = effort}
\DeclareAcronym{PE}{short = PE, long = performance}
\DeclareAcronym{FR}{short = FR, long = frustration}
\DeclareAcronym{SNR}{short = SNR, long = signal-to-noise ratio}
\DeclareAcronym{ANOVA}{short = ANOVA, long = Analyse of Variance}
\DeclareAcronym{LCD}{short = LCD, long = liquid crystal display}
\DeclareAcronym{ROS}{short = ROS, long = Robot Operating System}
\DeclareAcronym{FT}{short = F/T, long = force and torque, short-indefinite = an, long-indefinite = a}
\DeclareAcronym{BBT}{short = BBT, long = Box and Block Test}
\DeclareAcronym{ABBT}{short = ABBT, long = Aerial Box and Block Test}
\DeclareAcronym{MOCAP}{short = MOCAP, long = Motion Tracking System}

\renewcommand{\vec}[1]{\bm{#1}}		
\newcommand{\nR}[1]{\mathbb{R}^{#1}}		
\newcommand{\upperRomannumeral}[1]{\uppercase\expandafter{\romannumeral#1}}	

\renewcommand{\frame}[1]{\mathcal{F}_{#1}}		

\newcommand{\origin}{O}						
\newcommand{\vX}{\vec{x}}					

\newcommand{\vY}{\vec{y}}					
\newcommand{\vZ}{\vec{z}}					
\newcommand{\pos}{\vec{p}_B}				
\newcommand{\posRef}{\vec{p}_{B,\text{ref}}}    
\newcommand{\velRef}{\vec{v}_{B,\text{ref}}}	

\newcommand{\velMax}{{v}_{max}}

\newcommand{\frameW}{\frame{W}}			
\newcommand{\frameH}{\frame{H}}			
\newcommand{\frameC}{\frame{C}}			
\newcommand{\originW}{\origin_W}		
\newcommand{\originH}{\origin_H}		
\newcommand{\originC}{\origin_C}		
\newcommand{\xW}{\vX_W}				
\newcommand{\yW}{\vY_W}				
\newcommand{\zW}{\vZ_W}				
\newcommand{\xH}{\vX_H}				
\newcommand{\yH}{\vY_H}				
\newcommand{\zH}{\vZ_H}				
\newcommand{\xC}{\vX_C}				
\newcommand{\yC}{\vY_C}				
\newcommand{\zC}{\vZ_C}				

\newcommand{\angVel}{\dot{{\vec{\omega}}}}

\newcommand{\angRot}{\vec{\omega}}

\newcommand{\wrenchTotal}{\wrench_\text{fb,total}}
\newcommand{\wrenchRec}{\wrench_\text{fb,rec}}
\newcommand{\wrenchFbExt}{\wrench_\text{fb,ext}}

\newcommand{\wrench}{\bm{\tau}}

\def\BibTeX{{\rm B\kern-.05em{\sc i\kern-.025em b}\kern-.08em
    T\kern-.1667em\lower.7ex\hbox{E}\kern-.125emX}}
\begin{document}

\title{HATPIC: An Open-Source Single Axis \\ Haptic Joystick for Robotic Development \\
\thanks{This project has received funding from the European Union’s Horizon 2020 research and innovation program under the Marie Skłodowska-Curie grant agreement No 953454.}
}

\author{\IEEEauthorblockN{1\textsuperscript{st} Julien Mellet}
\IEEEauthorblockA{\textit{Dept. of Elec. Eng. and Info. Tech.} \\
\textit{University of Naples Federico II}\\
Naples, Italy \\
julien.mellet@unina.it}
\and
\IEEEauthorblockN{2\textsuperscript{nd} Fabio Ruggiero}
\IEEEauthorblockA{\textit{Dept. of Elec. Eng. and Info. Tech.} \\
\textit{University of Naples Federico II}\\
Naples, Italy \\
fabio.ruggiero@unina.it}
\and
\IEEEauthorblockN{3\textsuperscript{rd} Vincenzo Lippiello}
\IEEEauthorblockA{\textit{Dept. of Elec. Eng. and Info. Tech.} \\
\textit{University of Naples Federico II}\\
Naples, Italy \\
vincenzo.lippiello@unina.it}
}

\maketitle

\begin{abstract}
Humans process significantly more information through the sense of touch than through vision. Consequently, haptics for telemanipulation is poised to become essential in the coming years, as it offers operators an additional sensory channel crucial for interpretation in extreme conditions. However, current haptic device setups are either difficult to access or provide low-quality force feedback rendering. This work proposes the design of a single-axis, open-source setup for telemanipulation development, aimed at addressing these issues.
We first introduce the haptic device and demonstrate its integration with common robotic tools. The proposed joystick has the potential to accelerate the development and deployment of haptic technology in a wide range of robotics applications, enhancing operator feedback and control.
\end{abstract}
\vspace{0.2cm}


\section{Introduction}
For telepresence enhancement, haptics has some advantages over vision feedback.
While human vision processes motion at approximately $75Hz$~\cite{potter2014}, tactile feedback can be processed around $400Hz$~\cite{JOHANSSON198217}, allowing for quicker intuitive analysis. Various haptic devices, such as wearables~\cite{wearable-haptics}, robotic arms~\cite{mike-6dof}, and force-feedback joysticks~\cite{rotation-ctl}, have been used to provide force feedback. However, standard robotic joysticks often have a gamepad form factor~\cite{standard-joy}, where users control the vehicle with their fingers.
Additionally, haptic feedback is crucial for early-stage psychomotor skill acquisition~\cite{learning-surgery-robot}, making learning more effective thanks to haptic cues.

Given the advantages of haptics in \ac{HRI}, various devices have implemented it.
The primary open-source haptic device in related works is the Hapkit~\cite{hapkit}. While it was designed for educational purposes and has improved accessibility, it has several notable limitations. Despite enhancements in transmission, the capstan in the latest version remains difficult to assemble and is prone to friction slides, which can affect performance. Furthermore, the DC motor used in the Hapkit is fragile and provides insufficient force feedback, which is crucial for ensuring operator immersion and control during telemanipulation tasks.
Additionally, the joystick design requires unconventional handling, making it less intuitive for users. Another significant drawback is the lack of integration with standard communication frameworks like \ac{ROS}.

Reducing the number of controlled \ac{DoF} enhances steering accuracy and operator performance~\cite{rotation-ctl}. Consequently, a single-axis device is particularly well-suited for development of telemanipulation strategies. For steering across multiple dimensions, approaches such as shared autonomy and operator axis selection can be employed effectively.

In summary, proposing an open-source, single-axis haptic device would address an existing gap. The design of a universal force-feedback joystick for robotic applications is currently lacking in the literature.
Thus, the main contributions of this abstract are \begin{enumerate*}[label=\textit{\roman*)}]
    \item he hardware design of the device, and
    \item the presentation of the internal controller.
\end{enumerate*} 

\begin{figure}[t]
    \centering
    \includegraphics[width=1\linewidth]{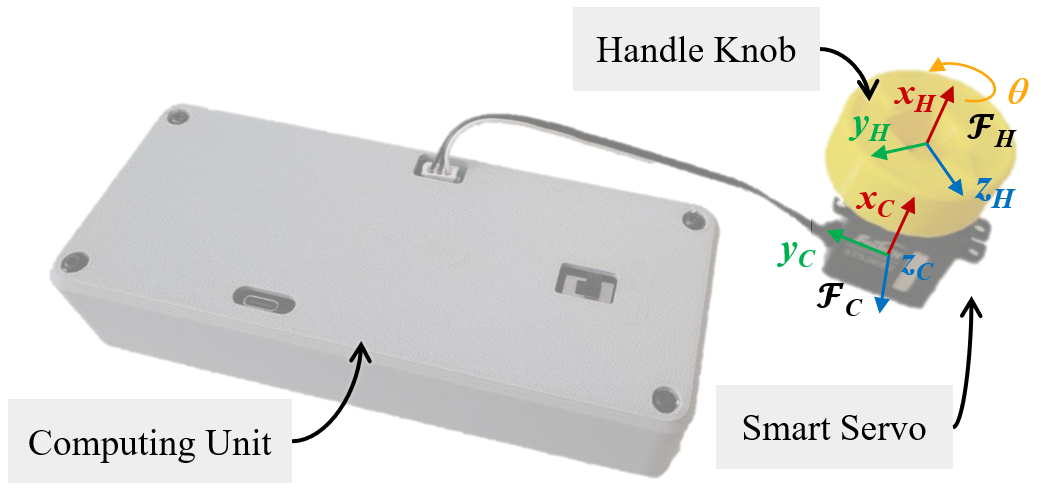}
    \caption{Components of the Hatpic joystick including the enclosed powerboard, the smart actuator with reference $\frameC$ and the handle knob with its frame $\frameH$.}
    \label{fig:eye-catcher}
    \vspace{-0.4cm}
\end{figure}

\section{Joystick Model}
The haptic device allows the operator to send reference to the robot. The inertial frame $\frameC=\{\originC,\xC,\yC,\zC\}$ with origin $\originC$ corresponds to the idle orientation of the sticks with $\zH$ colinear to $\zC$ (see Fig.~\ref{fig:eye-catcher}). The current orientation is described with the frame $\frameH=\{\originH,\xH,\yH,\zH\}$ with origins $\originH$ fixed with respect to the sticks. Orientation and angular rates of $\frameH$ with respect to $\frameC$ are respectively defined as $\theta, \angRot \in \nR{}$.

The robot is described by its position, $\pos \in \nR{3}$, relative to the inertial world frame $\frameW$. Here, $\originW$ denotes the origin, and ${\xW,\yW,\zW}$ are unit vectors, positioned at an arbitrary fixed point, with $\zW$ opposite to the direction of gravity.

Following \cite{mike-6dof}, we use an admittance filter with a low-level joint position controller for compliant interaction and haptic transparency. Assuming perfect joystick tracking, the closed-loop dynamics approximate as
\begin{equation}
    \label{haptic_model_ref}
    D_{adm} \; \angVel + M_{adm} \; \angRot = -\wrench + \wrenchTotal,
\end{equation}
where $ D_{adm}, M_{adm}  \in \nR{} $ represent the inertia and damping coefficient, respectively, both determined by user preferences, and $ \wrench, \wrenchTotal \in \nR{} $ are the interaction torque with the device and the total torque feedback applied to the operator, respectively. We define it as
\begin{equation}
    \label{eq:total-wrench}
    \wrenchTotal = \wrenchRec + \wrenchFbExt,
\end{equation}
where $\wrenchRec \in \nR{}$ is the re-centering torque and $\wrenchFbExt \in \nR{}$ is the interaction torque. In details,

\begin{equation}
    \label{recentering-wrench}
    \wrenchRec = -K_{rec}(\theta) \; \theta,
\end{equation}
where $K_{rec}(\theta)$ represents the re-centering gain. We define it dynamically and introduce a virtual notch to let the operator feel the idle position with more force as
\begin{equation}
    \label{dynamic-Krec}
    \begin{aligned}
    K_{rec}(\theta) = \begin{cases}
        0, & \text{if } |\theta - \theta_0| < Q_{dz}, \\
        K_{max}, & \text{if } Q_{dz} \leq |\theta - \theta_0| \leq N, \\
        K_{t}, & \text{otherwise},
    \end{cases}
    \end{aligned}
\end{equation}
where $K_{t} = K_{min} + \Delta K \left(1 - \frac{|\theta - \theta_0| - N}{N}\right)$ is an intermediate variable, $\theta_0 \in \nR{}$ is the center position, $ Q_{dz} \in \nR{+}$ is the deadzone range around the center where no torque is applied, $ N \in \nR{+}$ is the range where the notch effect is felt, and $ K_{min}, K_{max} \in \nR{+}$ are the minimum and maximum values for the stiffness.

The translational reference for the robot is given by
\begin{equation}
    \label{eq:ref-input}
    \begin{aligned}
        \velRef &= \frac{\velMax}{2} \ \theta, \quad
        \posRef = \int_{0}^{t} \velRef(b) \,db,
    \end{aligned}
\end{equation}
where $\velMax$ is the maximum velocity set by the operator preference. An equivalent formulation can be described for rotational reference.

\section{Design}
The actuation mechanism is a critical component of any haptic device. In our design, we have integrated a smart servo equipped with position and torque feedback. The device features custom electronics that allow powering by a battery and connect the smart servo to a microcontroller. This microcontroller, in turn, interfaces with a serial link via a USB-C port. The servo motor is capable of delivering up to $0.44$~Nm or $20$~N at the finger position, providing robust force feedback that enhances operator immersion and accommodates a wide range of force rendering applications.

The implementation of Eq. (\ref{eq:total-wrench}) operates as a separate thread on the device's microcontroller. A second thread is responsible for sending joystick position references via a tailored dataframe for serial communication. On the computer, the joystick driver functions as middleware, processing the device's dataframes and streaming them into the robot's network connection in either ROS1 or ROS2. The driver also connects the feedback data in a similar manner. The code for the framework, along with the open-source hardware, is available online\footnote{\url{https://github.com/jumellet/hatpic.git}}.

The smart servo is operated by the user with one hand holding the motor case while the other hand actuates the knob, which is directly attached to the actuator, simplifying the mechanism. The electronics is protected within a two-part enclosure casing. All components can be 3D printed, and the device is available in the repository.

The haptic joystick has demonstrated its robustness during use, with four units successfully manufactured. These devices were effectively employed to teach haptic bilateral teleoperation of aerial manipulators during the Summer School of the AERO-TRAIN European project.


\section{Conclusion}
To accelerate the development of haptics in robotic applications, we designed a single-axis joystick with force feedback control. Both the hardware and software are shared as open-source, including joystick drivers, electronics, and parts to be manufactured. The low-level force controller was described and successfully deployed in a robotic force application.

Future work will leverage the form factor to implement additional haptic rendering modalities. We plan to integrate virtual detents for precise position control and high-frequency oscillations for other rendering feedback information.

\bibliographystyle{IEEEtran}
\bibliography{bibliography.bib}

\end{document}